%% file: root.tex
\ifdefined\pdfminorversion
\fi
\RequirePackage[T1]{fontenc}
\documentclass[letterpaper, 10 pt, conference]{ieeeconf}  

\IEEEoverridecommandlockouts                              
\usepackage{graphicx}
\usepackage{newunicodechar}
\newunicodechar{’}{'}
\newunicodechar{—}{---}
\newunicodechar{×}{\ensuremath{\times}}
\usepackage{cite}
\usepackage{multicol}
\usepackage{multirow}
\usepackage{amsmath}
\usepackage{amssymb}
\usepackage{makecell}
\usepackage[table]{xcolor}
\usepackage{booktabs,tabularx}
\usepackage{pifont}
\usepackage{array}
\usepackage{float}
\makeatletter
\def\endfigure{\end@float}
\def\endtable{\end@float}
\makeatother
\usepackage[caption=false,font=footnotesize]{subfig}
\usepackage{capt-of}
\usepackage[pdfversion=1.4]{hyperref}
\ifdefined\XeTeXversion
  
\fi

\definecolor{blue}{HTML}{0173B2}
\definecolor{green}{HTML}{06A66C}
\definecolor{red}{HTML}{CF3C33}
\definecolor{magenta}{HTML}{D748A2}
\definecolor{purple}{HTML}{A327F5}
\definecolor{pink}{HTML}{D45990}
\definecolor{BlueGreen}{HTML}{088F8F}

\newcommand{\cmark}{{\color{green}\ding{51}}}  
\newcommand{\xmark}{{\color{red}\ding{55}}}    

\newcolumntype{Y}{>{\centering\arraybackslash}X}
\newcolumntype{C}[1]{>{\centering\arraybackslash}p{#1}} 
\newcolumntype{L}[1]{>{\raggedright\arraybackslash}p{#1}} 

\newcommand{\xxnote}[3]{}
\ifx\hidenotes\undefined
  \renewcommand{\xxnote}[3]{\color{#2}{#1: #3}}
\fi

\hypersetup{
    hidelinks,
}

\title{\LARGE \bf
\raisebox{-0.2em}{\includegraphics[height=1.1em]{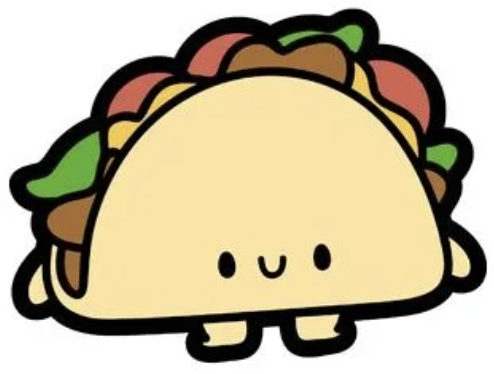}}~TacO: Benchmarking Tactile Sensors\\for Object Manipulation
}

\newcommand{\tacoauthor}[2]{%
  \begin{tabular}[t]{@{}c@{}}\textbf{#1}\\[-1pt]#2\end{tabular}%
}
\author{%
\begin{minipage}{\textwidth}
\centering\normalsize
\begin{tabular}{@{}c@{\hspace{1.8em}}c@{\hspace{1.8em}}c@{\hspace{1.8em}}c@{}}
\tacoauthor{Anya Zorin}{UC San Diego} &
\tacoauthor{Zilin Si}{CMU} &
\tacoauthor{Sachin Bhadang}{UC San Diego} &
\tacoauthor{Myungsun Park}{UC San Diego}
\end{tabular}\endgraf\vspace{1em}
\begin{tabular}{@{}c@{\hspace{1.5em}}c@{\hspace{1.5em}}c@{\hspace{1.5em}}c@{\hspace{1.5em}}c@{}}
\tacoauthor{Junsung Park}{UC San Diego} &
\tacoauthor{Alexiy Buynitsky}{UC San Diego} &
\tacoauthor{Taejun Park}{SNU} &
\tacoauthor{Sohee John Yoon}{SNU} &
\tacoauthor{Yong-Lae Park}{SNU}
\end{tabular}\endgraf\vspace{1em}
\begin{tabular}{@{}c@{\hspace{1.5em}}c@{\hspace{1.5em}}c@{\hspace{1.5em}}c@{\hspace{1.5em}}c@{}}
\tacoauthor{Oliver Kroemer}{CMU} &
\tacoauthor{Zeynep Temel}{CMU} &
\tacoauthor{Michael T. Tolley}{UC San Diego} &
\tacoauthor{Sha Yi}{UC San Diego} &
\tacoauthor{Xiaolong Wang}{UC San Diego}
\end{tabular}
\end{minipage}%
}

\IEEEaftertitletext{\input{figures/sensor_fig}\endgraf\vspace{\baselineskip}}

\begin{document}

\maketitle
\thispagestyle{empty}
\pagestyle{empty}
\raggedbottom

\input{0_abstract}


\input{1_introduction}
\input{2_related_work}
\input{3_tactile_sensors}

\input{4_policy_training}
\input{5_experiments}

\input{6_results}
\clearpage
\input{7_discussion}

\input{acknowledgments}

\bibliographystyle{IEEEtran}
\bibliography{ref}

\onecolumn
\input{a_appendix}

\end{document}

%% file: figures/sensor_fig.tex
\begin{minipage}{\textwidth}
    \centering
    \includegraphics[width=0.92\textwidth]{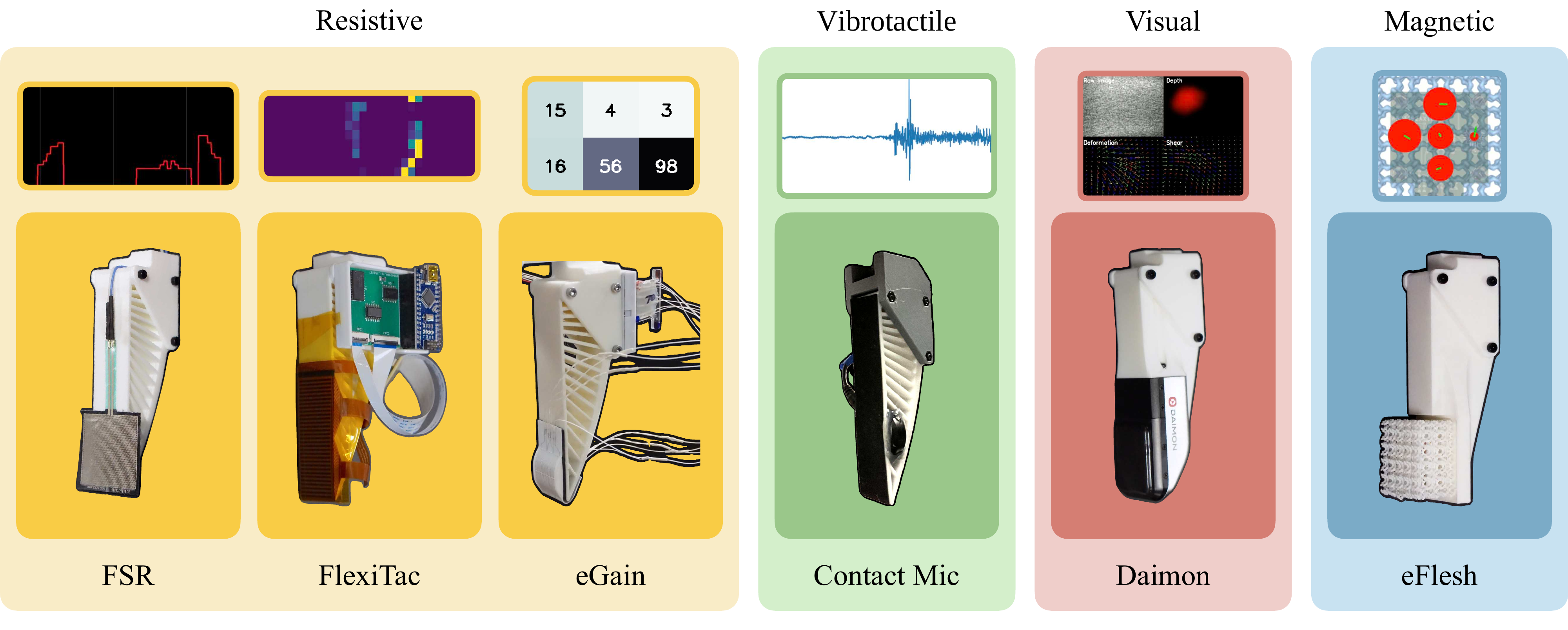}
    \captionof{figure}{The six sensors evaluated in TacO grouped by sensing modality, with example sensor readings.}
    \label{fig:sensor_types}
\end{minipage}

%% file: 0_abstract.tex
\begin{abstract}
Vision-based learning from demonstrations has achieved remarkable success in enabling robots to perform manipulation tasks and high-level semantic reasoning, yet it remains insufficient for complex, contact-rich manipulation. While there is broad agreement that tactile sensing improves manipulation, there is no empirical guidance on which tactile sensors are best suited for which manipulation tasks. In this paper, we provide a systematic, task-driven evaluation of tactile sensors for robot manipulation and propose a framework for selecting and evaluating sensors based on manipulation policy performance. Separate manipulation policies are trained for tactile sensors of four distinct modalities: visual, acoustic, magnetic, and resistive, across three tasks: pick-and-place with unknown mass, object reorientation, and plug insertion. For each task, we analyze how sensor properties, including spatial resolution, shear sensing, tactile representation, and inherent material friction, affect task performance. Rather than tactile sensing being universally beneficial in the same way, our results show that the usefulness of tactile information depends strongly on sensor modality, material properties, and the specific manipulation tasks. The tactile sensors, code, data, and hardware setup will be publicly available on the project website.
\end{abstract}

%% file: 1_introduction.tex
\section{Introduction}

Tactile sensing is a critical component of contact-rich manipulation. While recent advances in learning-based robotics have enabled impressive manipulation capabilities using vision alone, visual feedback lacks access to many of the physical interaction signals needed for task success. A broad consensus has emerged that tactile sensing can improve object manipulation performance \cite{calandra2018more, qi2023general, huang2025vt}. Despite the wide availability of tactile sensors, there remains little empirical guidance on how different tactile sensing modalities affect learned manipulation policies or which sensors are most appropriate for which tasks.

\input{tables/benchmark_compare}

Existing tactile benchmarks primarily focus on evaluating sensor hardware in isolation through calibration, characterization, or perceptual tasks~\cite{gao2022objectfolder, luu2025manifeel}, and those that do incorporate policy learning tend to use high-resolution visual tactile sensors on object reconstruction or pattern recognition tasks~\cite{schneider2025tactile}. When evaluating tactile sensors, it is difficult to determine whether performance gains arise from the sensing modality itself, the task design, or the learning pipeline. This makes it challenging for researchers to choose tactile sensors when designing manipulation systems, especially when balancing performance, cost, accessibility, and ease of integration.

In this work, we aim to holistically evaluate tactile sensors for object manipulation using an imitation learning pipeline. We adopt Action Chunking Transformers (ACT)~\cite{zhao2023learning, qiu2025humanoid} as our policy backbone and build a modular pipeline that supports a variety of tactile encoders. ACT predicts actions in fixed-length chunks using a CVAE architecture, capturing multimodal demonstration distributions, and its transformer-based encoder accommodates additional input modalities as tokens, making it well-suited for integrating diverse tactile inputs. Tactile encoding methods are sensor-specific and include MLP encoding of raw sensor values, convolutional encoding of tactile images, and spectral processing of vibrotactile audio.

Multiple policies are trained per sensor across three representative manipulation tasks: pick-and-place with unknown object mass, plug insertion with occluded contact geometry, and object reorientation requiring continuous force modulation. Both vision-only and visuotactile policies are trained on the same data, with tactile readings removed for the vision-only case.

With this pipeline, we evaluate six tactile sensors spanning four modalities: resistive, magnetic, vision-based, and vibrotactile, covering a wide range of data dimensionalities, costs, and material properties (Fig.~\ref{fig:sensor_types}). By including readily available tactile sensors, we lower the barrier to entry for researchers integrating tactile sensing into manipulation pipelines. To ensure reproducibility, we evaluate this pipeline across two institutions located over 2,000 miles apart using comparable robot platforms and independent data collection setups.

Tactile sensors are the direct interface between robot and object: surface materials that mediate contact also drive the sensing signal, making embodiment and sensing fundamentally coupled. We design two analyses to control for each factor independently: a per-sensor analysis comparing vision-only and visuotactile policies to isolate the contribution of tactile signals, and a cross-sensor vision-only analysis to isolate embodiment effects, where performance differences reflect sensor material, appearance, and form factor rather than sensing capability. We additionally conduct a hardware repeatability analysis using an open-source test kit. 
The main contributions of this paper are:
a) An imitation learning pipeline and real-world evaluation of six tactile sensors across three tasks;
b) Per-sensor analysis of tactile contribution and cross-sensor analysis of embodiment effects;
c) Hardware repeatability analysis and open-source test kit built from off-the-shelf components.

%% file: tables/benchmark_compare.tex
\begin{table*}[!t]
\centering
\small
\setlength{\tabcolsep}{5pt}
\caption{TacO covers more tactile modalities and sensors than prior benchmarks, while providing a fully real-world, open-source evaluation across three manipulation tasks.}
\label{tab:tactile_benchmarks}
\rowcolors{3}{gray!10}{white}
\begin{tabular}{l >{\raggedright\arraybackslash}m{2.6cm} c c >{\raggedright\arraybackslash}m{4.5cm}}
\toprule
\textbf{Benchmark} &
\textbf{Modality} &
\textbf{\# Sensors} &
\textbf{\makecell{Open-source\\test kit}} &
\textbf{Tasks} 
 \\
\midrule
ObjectFolder~\cite{gao2022objectfolder, gao2023objectfolder} & Visual & 2 & \xmark & Reconstruction, Localization\\
Tactile MNIST~\cite{schneider2025tactile} & Visual & 1 & \xmark & Classification, Volume Estimation \\
VTDexManip~\cite{liuvtdexmanip} & Resistive & 1 & \cmark & 6 Dexterous Manipulation Tasks  \\
ManiFeel~\cite{luu2025manifeel} & Visual & 1 & \cmark & Insertion, Screwing, Exploration \\
\textbf{TacO (Ours)} & Vibrotactile, Magnetic, Resistive, Visual & 6 & \cmark & Pick \& Place, Insertion, Reorientation \\
\bottomrule
\end{tabular}
\end{table*}

%% file: 2_related_work.tex
\section{Related Work}
\input{tables/sensor_compare}
Recent progress in imitation learning (IL) has made it possible to train manipulation policies directly from demonstrations~\cite{chi2023diffusion, zhao2023learning, fu2024-mobilealoha}. Teleoperation has become a dominant way to collect high-quality manipulation demonstrations at scale~\cite{yang2024ace, cheng2024open, wu2023gello}. Despite these advances, most widely used IL pipelines remain vision-dominant~\cite{black2024-pi0, bjorck2025gr00t, lee2025molmoact}. This is especially limiting for contact-rich manipulation tasks, where the forces and contact events heavily influence the interaction outcome, motivating policies that can leverage tactile sensor information alongside images and videos. 

Tactile sensors offer observations of contact geometry, pressure, shear, and high-frequency interactions that vision often overlooks~\cite{kappassov2015tactile}. Existing tactile sensors span diverse modalities with different trade-offs in spatial resolution, bandwidth, robustness, and cost. Vision-based tactile sensors offer dense image measurements of deformation~\cite{tremblay1993estimating, yuan2017gelsight, lambeta2020digit, li2019elastomer, sferrazza2019design, lambeta2024digit360}. Magnetic tactile sensors infer contact through magnet motion in a soft medium, often providing low-latency, compact sensing with sensitivity to shear and normal forces~\cite{tomo2018new, hellebrekers2019soft, dai2022design, bhirangi2025anyskin, pattabiraman2025eflesh}. Resistive tactile sensors~\cite{park2012design, bhattacharjee2013tactile, yin2023rotating, gupta2023forcesticker, huang20243d, yoshimura2025m3d} provide direct readings with simpler hardware but can face challenges in hysteresis, drift, and spatial sparsity depending on the layout. Contact microphones capture vibration and acoustic signals~\cite{arandjelovic2017look, zhang2019leveraging, mejia2024hearing, aderibigbe2025milli, xu2025multi} from frictional events and impacts, providing high temporal bandwidth information for slip and texture even when spatial information is limited. 
While many prior systems demonstrate strong results with one modality, differences in embodiments, tasks, and learning pipelines make it difficult to attribute gains to the sensing modality itself.

Benchmarking in robotics is challenging due to hardware variations and inconsistent real-world evaluations. Simulation-based evaluations~\cite{si2022taxim, huang2025vt} have played a valuable role because they are easy to distribute and run at scale. However, it is challenging to simulate a wide range of tactile sensor readings with high fidelity. Additionally, for contact-rich manipulation, simulation does not reliably represent real-world behaviors. Most previous benchmarks focus on a single sensing modality [Table \ref{tab:tactile_benchmarks}] ~\cite{gao2022objectfolder, gao2023objectfolder, schneider2025tactile, liuvtdexmanip, luu2025manifeel}. An increasing number of real-world datasets include tactile measurements across a variety of sensors to incorporate more modalities~\cite{song2025opentouch, higuera2025tactile}. While vision-only models have achieved substantial progress without contact information, it is natural to raise a practical question: when is tactile sensing actually beneficial and which tactile signals are most useful for policy learning? To address this gap, we present a controlled, real-world benchmark that compares tactile modalities across tasks using the same teleoperation and policy training framework. We identify which modalities provide measurable gains, under what conditions, and how they complement vision-only policies.

%% file: tables/sensor_compare.tex
\begin{table*}[!t]
\centering
\small
\setlength{\tabcolsep}{4pt}
\caption{Comparison of tactile sensors benchmarked for robotic manipulation. For FSR and eGain, readings are obtained on demand (OD) at a frequency determined by external devices.}
\label{tab:tactile_sensor_compare}
\rowcolors{2}{white}{gray!10}
\begin{tabularx}{\linewidth}{
  L{2.65cm}   
  *{6}{Y}    
}
\toprule
\textbf{} &
\textbf{FSR} &
\textbf{FlexiTac} &
\textbf{eGain} &
\textbf{eFlesh} &
\textbf{Daimon} &
\textbf{Contact Mic} \\
\midrule
Modality           & Resistive & Resistive   & Resistive   & Magnetic    & Visual        & Vibrotactile \\
Normal Force (N)   & 0.2$-$20.0 & 0.2$-$10.0 & 0.0$-$27.5 & 0.0$-$30.0 & 0.3$-$30.0   & $-$ \\
Shear Force (N)    & $-$       & $-$         & $-$         & 0.0$-$17.5  & 0.1$-$8.0     & $-$ \\
Taxel Size (mm)    & 34$\times$34 & 2$\times$2 & 8.4$\times$12.7 & 0.5$\times$0.5 & 0.1$\times$0.05 & 30$\times$30 \\
Spatial Res.\ (px) & 1         & 12$\times$32 & 2$\times$3 & 5           & 320$\times$240 & 1 \\
Frequency (Hz)     & OD        & 20$-$100    & OD          & 100         & 60 / 120      & 44100 \\
Response Time (s)  & 0.017$\pm$0.026 & 0.174$\pm$0.063 & 1.132$\pm$0.581 & 0.029$\pm$0.024 & 0.132$\pm$0.038 & 0.037$\pm$0.020 \\
Std                & 0.002     & 0.063       & 0.125       & 0.047       & 0.036         & $7e^{-4}$ \\
High Friction?     & \smash{\xmark} & \smash{\xmark} & \smash{\cmark} & \smash{\cmark} & \smash{\cmark} & \smash{\cmark} \\
Price (USD)        & \$5       & \$35        & \$5         & \$35        & \$965         & \$27 \\
\bottomrule
\end{tabularx}
\label{tab:sensor-character}
\end{table*}

%% file: 3_tactile_sensors.tex

\section{Tactile Sensors Overview}
\label{sec:sec_3}

Tactile sensors are designed to capture physical interaction cues that are difficult or impossible to infer from vision alone, such as contact forces, deformation, slip, and vibration. These design choices lead to different trade-offs in spatial resolution, force sensitivity, repeatability, ease of integration, form factor, and sensor material. As a result, different manipulation tasks benefit from different combinations of tactile sensing capabilities. 
A comparison of sensors is provided in Table~\ref{tab:tactile_sensor_compare}.

\textbf{Piezoresistive Sensors:}
Piezoresistive tactile sensors measure contact through changes in electrical resistance induced by deformation. They are widely used due to their simplicity, low cost, and ease of integration, but typically provide limited access to shear information and may exhibit hysteresis or sensor reading drift. A \textit{Force Sensing Resistor (FSR)} is an off-the-shelf resistive component that outputs a single scalar value proportional to applied normal force. It is thin, inexpensive, and can be mounted directly on robotic grippers with tape or glue. \textit{FlexiTac} \cite{huang20243d}  is an open-source resistive tactile sensor consisting of a 12×32 array of taxels that measure spatially distributed normal force. The sensor is fabricated using layered elastomeric materials with embedded conductive traces, resulting in a compliant sensing surface. A \textit{Liquid Metal Resistive Sensor (eGain)} \cite{park2012design} is a resistive tactile sensor composed of microchannels embedded in an elastomeric substrate (e.g., silicone) and filled with eutectic gallium-indium. The sensor has to be custom-made either by hand or by a specialized instrument that prints liquid metal traces on silicone. When pressure is applied, the elastomer deforms, changing the geometry of the channel and increasing the electrical resistance. The sensor's sensitivity and spatial resolution depend on the microchannel design, while the elastomer and liquid metal provide compliance and flexibility.

\textbf{Vibrotactile Sensors:} A \textit{Contact Microphone}  is a low-cost vibrotactile sensor constructed from a piezoelectric disc mounted beneath the fingertip contact surface. It outputs a single-channel, high-frequency signal capturing vibrations generated during contact events such as impacts, slip, and sliding. While it provides no spatial information or direct force measurement, it offers high temporal bandwidth and is mechanically decoupled from surface wear, which can improve robustness.

\textbf{Vision-based Tactile Sensors:} \textit{Daimon}\footnote{\url{https://www.dmrobot.com/en/product/p1/dm-tac-w.html}} uses a soft membrane along with a camera. The camera captures the deformation of this membrane and translates it into the forces applied based on material properties. Vision-based sensors offer significantly higher spatial resolution compared to other modalities. However, the rigid camera-based structure also imposes constraints on form factor.

\textbf{Magnetic Sensors:} \textit{eFlesh} ~\cite{pattabiraman2025eflesh} is an open-source, 3D-printed magnetic tactile sensor that infers contact forces from the displacement of embedded magnets measured by Hall-effect sensors. Each sensing unit provides a three-dimensional force vector, enabling estimation of both normal and shear forces, but offers lower spatial resolution than dense resistive or visual tactile sensors.

\begin{figure*}[!t]
\centering
\includegraphics[width=\linewidth]{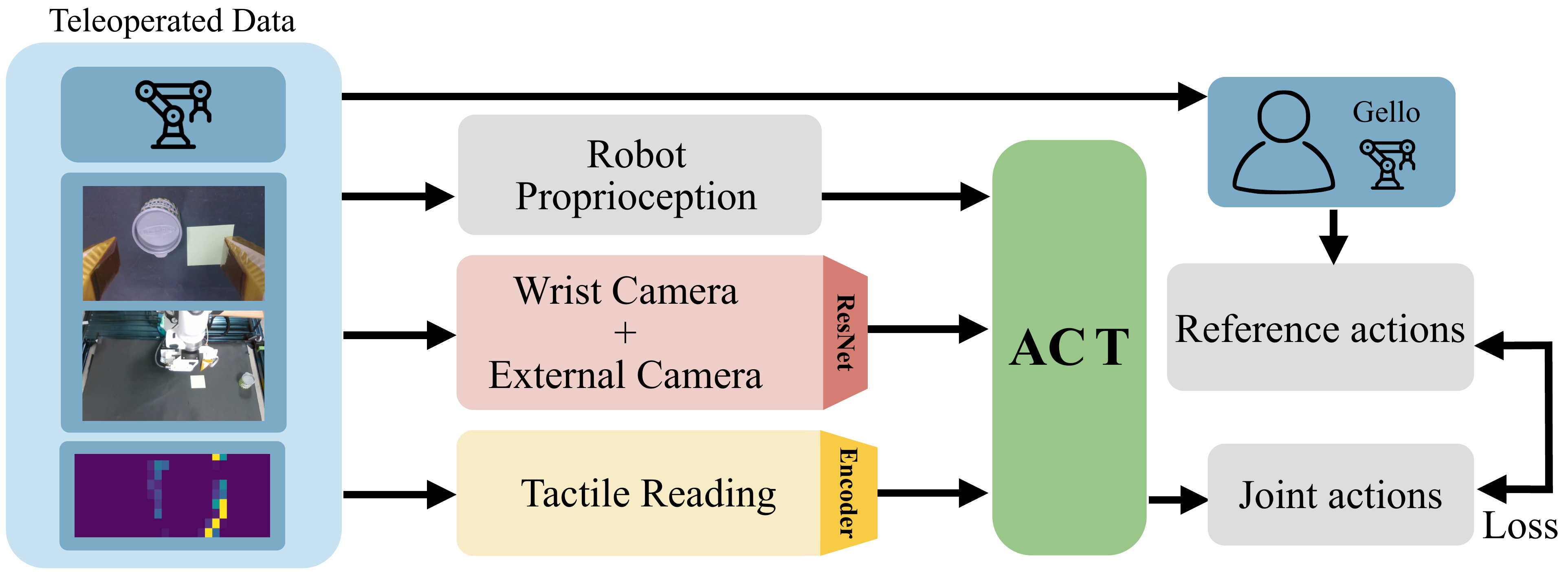}
\caption{TacO imitation learning pipeline with tactile, camera, and proprioceptive observations as inputs.}
\label{fig:framework}
\end{figure*}

%% file: 4_policy_training.tex
\section{Policy Training}
\label{sec:sec_4}
As shown in Figure~\ref{fig:framework}, we train visuomotor policies with Action Chunking Transformers (ACT)~\cite{zhao2023learning, qiu2025humanoid} that are conditioned on different tactile sensor modalities. We collect demonstrations from teleoperation using GELLO~\cite{wu2023gello} or Factr~\cite{liu2025factr}. Each demonstration provides a time-indexed sequence of observations and actions $\{(o_t, a_t)\}_{t=1}^T$, where $o_t$ includes (i) RGB images $I_t$ from one wrist-mounted camera and one third-person camera, (ii) robot proprioception $p_t$ consisting of gripper width and joint states, and (iii) tactile measurements for modality $m \in \{\text{visual}, \text{magnetic}, \text{resistive}, \text{vibrotactile}\}$.

Following action-chunked behavior cloning, the policy $\pi$ predicts a fixed horizon-$H$ action chunk $\hat{a}_{t:t+H}$ and executes it in a receding-horizon manner. We train two independent policies for each task and sensor configuration: a visuotactile policy and a vision-only policy. They are trained on the same data, with the tactile data removed for the vision-only policy. 

\textbf{Tactile Encoders:}
\label{sec:sec_4.1}
The tactile sensors differ substantially in signal type and dimensionality, so tactile data from each sensor are handled in a modality-specific manner. We experiment with both raw sensor readings and signals preprocessed into more physically meaningful representations, as well as various encoding methods. All tactile encodings are concatenated with visual and proprioceptive features prior to policy learning. \textit{Raw Tactile Data:} For the FSR, FlexiTac, eGain, and eFlesh sensors, we train models using raw sensor readings. Since the FSR outputs a single scalar value, this value is linearly projected to $\mathbb{R}^{512}$. FlexiTac, eGain, and eFlesh produce higher-dimensional signals, which are encoded using an MLP. \textit{Tactile Images:}
The Daimon sensor is a vision-based tactile sensor that outputs raw RGB images, as well as processed depth, deformation, and shear images. For the FlexiTac sensor, we convert the raw array of values into a tactile image. We use all of these images as inputs and encode them using principal component analysis (PCA) or ResNet18 \cite{he2016deep}. \textit{Vibrotactile:} The contact microphone captures high-frequency vibrotactile signals. During data collection, audio is streamed at 44.1~kHz and converted from the time domain to a mel-spectrogram representation using a short-time Fourier transform (STFT) with a 0.5~s sliding window. The resulting mel-spectrogram has dimensions $128 \times 87$ and is flattened into an 11,136-dimensional feature vector, which is then encoded with an MLP.

\begin{figure*}[!t]
    \centering
    \includegraphics[width=\textwidth]{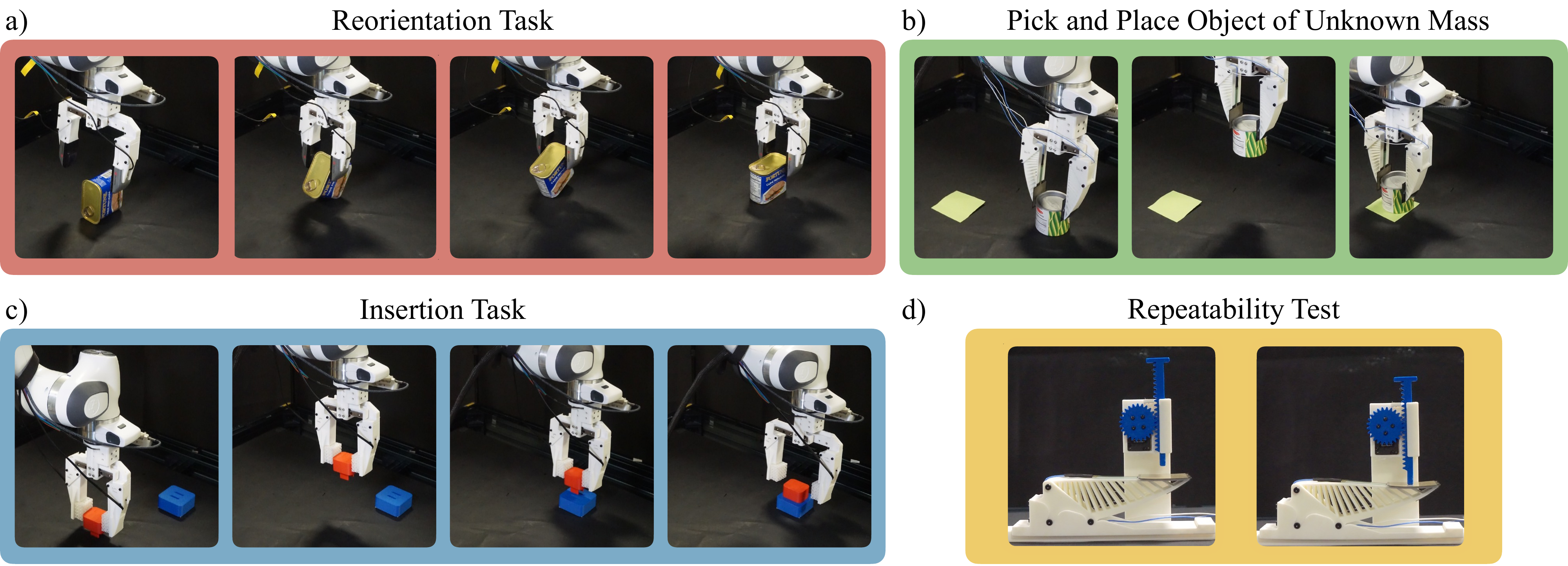}
    \vspace{-5pt}
    \caption{
    (a)--(c) Successful executions of the three manipulation tasks. (d) Sensor repeatability test setup.
    }
    \label{fig:plug_insertion}
\end{figure*}

\subsection{Training Pipeline}

We encode RGB observations using ResNet18~\cite{he2016deep} initialized with ImageNet~\cite{deng2009imagenet} weights and trained end-to-end to project images to $z_t^{\text{image}} \in \mathbb{R}^{512}$. This follows recent imitation learning work that leverages strong pretrained visual representations for improved robustness~\cite{qiu2025humanoid, huang2025vt}. The proprioceptive representation of joint angles, $p_t \in \mathbb{R}^{128}$, is linearly projected to $z_t^{\text{prop}} \in \mathbb{R}^{512}$ to match the transformer's hidden dimension. Tactile observations are encoded into modality-specific features $z_t^{\text{tactile}} \in \mathbb{R}^{512}$ as described in Section \ref{sec:sec_4.1}. For Daimon, the image-based tactile sensor, tactile features are concatenated spatially with camera features along the width dimension before being flattened into spatial tokens. For array-based tactile sensors (FSR, FlexiTac, eFlesh), each sensor's encoded features form individual tokens. Each token $x_t$ input to the transformer is a concatenation of the features mentioned above: $ x_t = [z_t^{\text{latent}}; z_t^{\text{prop}}; z_t^{\text{tactile}}; z_t^{\text{image}}]$,
where the latent token $z_t^{\text{latent}}$ is sampled from a learned prior during training and set to zero during inference. This sequence is processed by a transformer-based policy $f_\theta(\cdot)$~\cite{vaswani2017attention} with 4 encoder layers and 7 decoder layers (hidden dimension 512, 8 attention heads), which outputs an action chunk $\hat{a}_{t:t+H-1}$ where $H=64$. We train policies using behavior cloning with a CVAE objective that combines reconstruction and regularization losses:
\begin{equation}
\mathcal{L} = \sum_{\tau=0}^{H-1}\|\hat{a}_{t+\tau}-a_{t+\tau}\|_1 + \lambda_{\text{KL}} \cdot D_{\text{KL}}(q(z|a) \| p(z))
\end{equation}
where $q(z \mid a)$ is the CVAE encoder posterior and $p(z) = \mathcal{N}(0, I)$ is the standard normal prior, and $\lambda_{\text{KL}} = 10$. We optimize using the AdamW optimizer~\cite{loshchilov2019AdamW} with a learning rate $10^{-5}$ for both the policy and the ResNet18 backbone. During training, we apply image augmentations including random cropping (to 95\%, followed by resizing), random rotation ($\pm5^{\circ}$), and color jitter (brightness, contrast, and saturation variation of 0.3). We perform per-modality Gaussian normalization for all inputs (images, proprioception, tactile signals, and actions) using statistics computed from the training data. To prevent overfitting on proprioception, we apply conditional masking that randomly zeros out proprioceptive inputs during training. The policy predicts action chunks of length 64. At deployment time, we query the policy for 64-step chunks but execute only the first 32 steps in a receding-horizon manner before requerying, following standard practice in action-chunked policies~\cite{zhao2023learning}. At each 20~Hz policy step, we use the latest available timestamped tactile observation.

%% file: 5_experiments.tex
\section{Experiments}
\label{sec:sec_5}

\textbf{General Experimental Setup:}
Experiments were conducted across two institutions, both using Franka Panda 7-DOF arms. The first institution tested FSR, FlexiTac, eGain, Daimon, and eFlesh using the GELLO~\cite{wu2023gello} teleoperation system with a custom motorized gripper and finray~\cite{chi2024UMI} fingers, modified to accommodate each sensor as shown in Fig.~\ref{fig:sensor_types}. The second institution tested the contact microphone using the Factr~\cite{liu2025factr} teleoperation system with the standard Franka gripper and finray fingers. Both setups include a wrist-mounted RGB camera and a top-down RGB camera, with end-effector position control via the franky library~\cite{Schneider_franky_High-Level_Control}.

\textbf{Pick-and-Place with Unknown Object Mass:}
The pick-and-place task involves grasping a visually identical object with unknown internal mass. A can is used, which is empty in half of the trials and weighted with marbles in the other half. We collect 60 demonstrations per sensor (30 per mass condition), with the object's initial location randomized and the target location fixed. Success requires grasping the object without slipping and placing it accurately at the target. Each policy is evaluated over 20 rollouts, with 10 per mass condition.

\input{tables/pnp_mass_results}
\input{tables/reorient_results}

\textbf{Plug Insertion:}
The plug insertion task uses a 3D-printed plug and socket to evaluate whether tactile cues beyond normal force, such as shear force or vibration, improve insertion under partial visual occlusion. The socket location is randomized while the plug starts from a fixed position. The prongs are fully occluded during insertion, making tactile feedback critical for determining contact state. Success is defined as fully inserting the plug; partial success is recorded when the prongs are inserted only halfway. We select FlexiTac for spatial resolution, eFlesh for shear sensing, and the contact microphone for contact-event vibrations transmitted through the grasped object. We collect 50 demonstrations per sensor and evaluate each policy over 10 rollouts.

\textbf{Object Reorientation:}
The object reorientation task involves rotating a container on the table around a fixed point while maintaining contact with the surface, requiring continuous force modulation throughout. Success requires completing the reorientation without lifting or sliding the object off the table. We select Daimon for spatial resolution, eFlesh for shear sensing, and FSR for scalar normal-force measurements. We collect 50 demonstrations per sensor and evaluate each policy over 10 rollouts.

\textbf{Repeatability Test:}
We develop a low-cost test kit to evaluate sensor repeatability, using one Dynamixel motor and 3D-printed parts. A fingertip fixture holds the tactile sensor facing an indenter, which is actuated by a motor-driven rack-and-pinion system to apply a constant force. We collect 60 episodes over one hour, with each episode consisting of 30 seconds of pressing and 30 seconds of release. Since contact microphones only capture dynamic contact events, their repeatability test uses 60 shorter episodes of 6 seconds each (3 seconds of pressing), completed in 6 minutes. We report the time for each sensor to reach 90\% of its maximum reading and document the standard deviation across episodes to characterize hysteresis.

%% file: tables/pnp_mass_results.tex
\begin{table*}[tb]
\centering
\small
\caption{Pick-and-place successes for light and heavy objects}
\label{tab:pnp_mass}
\begin{tabularx}{\textwidth}{@{}l l *{6}{Y}@{}}
\toprule
\textbf{Method} & \textbf{Mass} & \textbf{FSR} & \textbf{FlexiTac} & \textbf{eGain} & \textbf{Contact Mic} & \textbf{Daimon} & \textbf{eFlesh} \\
\midrule
\multirow{2}{*}{Vision only} & Light & 0.6 & 0.6 & 0.6 & 0.7 & 1.0 & 0.9 \\
 & Heavy & 0.4 & 0.8 & 0.4 & 0.6 & 0.9 & 0.8 \\
\midrule
\multirow{2}{*}{Vision + tactile} & Light & 0.7 & 0.8 & 0.9 & 0.8 & 0.8 & 0.8 \\
 & Heavy & 0.3 & 0.9 & 0.6 & 1.0 & 0.8 & 1.0 \\
\bottomrule
\end{tabularx}
\end{table*}

%% file: tables/reorient_results.tex
\begin{table*}[tb]
\centering
\small
\caption{Success rates for plug insertion and object reorientation tasks.}
\begin{tabularx}{\linewidth}{l Y >{\centering\arraybackslash}p{2.0cm} Y c Y Y Y}
& \multicolumn{3}{c}{\textbf{Plug Insertion}} & & \multicolumn{3}{c}{\textbf{Object Reorientation}} \\
\cmidrule(lr){2-4} \cmidrule(lr){6-8}
\textbf{Method} & \textbf{FlexiTac} & \textbf{Contact Mic} & \textbf{eFlesh} & & \textbf{FSR} & \textbf{Daimon} & \textbf{eFlesh} \\
\midrule
Vision Only      & 0.1 & 0.2 & 0.3 & & 0.6 & 0.2 & 0.5 \\
Vision + Tactile & 0.3 & 0.7 & 0.7 & & 0.8 & 0.7 & 0.8 \\
\bottomrule
\end{tabularx}
\label{tab:insertion_reorient_results}
\end{table*}

%% file: 6_results.tex
\section{Results}
\label{sec:sec_6}

\subsection{Intra-Task}
\textbf{Pick-and-Place:} Overall, most visuotactile policies outperform their corresponding vision-only baselines [Tab.~\ref{tab:pnp_mass}], although the improvement is minor for many sensors. The Daimon tactile policy underperforms the vision-only policy. For FSR, vision-only policies typically fail to fully lift the heavy object, whereas tactile policies typically drop it halfway to the endpoint because the object slips on the low-friction material. The Daimon tactile policy exhibits failure modes involving collisions with the table, which are discussed further in Section~\ref{sec:sec_6}.
Tactile sensing improves performance primarily because vision alone cannot disambiguate object mass. Analysis of commanded gripper width shows that vision-only policies tend to output an average gripper width, whereas tactile policies learn distinct widths for light versus heavy objects. This indicates that tactile sensing enables the policy to infer latent physical properties and adapt its control accordingly.

\textbf{Plug Insertion:} All tactile policies show a substantial improvement over vision-only baselines, with significantly larger gains observed for eFlesh and the contact microphone as reported in Table~\ref{tab:insertion_reorient_results}.
In this task, vision and proprioception are primarily used for grasping and coarse alignment, while tactile sensing is critical during the final insertion phase. The failure mode of vision-only policies is placing the plug incorrectly, applying a small downward force, and then releasing the plug. In contrast, tactile policies slide the plug along the socket surface and often partially insert it before terminating, a failure mode that accounts for about 55\% of the tactile policy failures for FlexiTac. When insertion fails, tactile policies continue pushing downward and become stuck due to the lack of recovery demonstrations. These behaviors suggest that tactile information provides critical cues for detecting contact state and guiding insertion.

\textbf{Object Reorientation:}
All tactile policies outperform their vision-only counterparts [Tab.~\ref{tab:insertion_reorient_results}]. Daimon sees the largest relative improvement, followed by eFlesh and FSR. The FSR vision-only policy achieves a higher baseline than the other sensors, which we attribute to its low-friction surface; the high-friction compliant sensors (eFlesh and Daimon) share a failure mode in which the object is lifted but fails to rotate against the table, or is dropped without rotation. Despite Daimon's higher spatial resolution and richer shear information compared to eFlesh, all tactile policies achieve similar success rates of around 80\%. This suggests that for continuous manipulation tasks requiring force regulation, access to appropriate tactile feedback is more important than high spatial resolution or complex tactile representations. Overall, the results indicate that simpler, lower-cost tactile sensors can be sufficient for fine-grained manipulation when paired with an appropriate learning pipeline.

We trained separate reorientation policies using normal-only and shear-only inputs for Daimon and eFlesh (Table~\ref{tab:force_components}), as these are the only two sensors in our evaluation that provide shear-sensing channels. The relative contribution of each force component varies between sensors.
\input{tables/force_components}

\input{tables/sensor_material}

\input{tables/expanded_repeatability}
\input{figures/repeatability_results}

\subsection{Cross-Task}
\textbf{Sensor Material and Form Factor:}
\label{sec:sensor_material} Sensor material and surface properties have a clear impact on manipulation performance. The FSR and FlexiTac sensors present relatively low-friction, slippery contact surfaces, whereas eFlesh, Daimon, and the contact microphone are integrated with more compliant, higher-friction materials.
To isolate the impact of this, we compare the performance of \textbf{vision-only} policies across sensors and tasks, where tactile signals are not used. Vision-only policies using high-friction, compliant sensors achieve higher average success rates across tasks [Tab.~\ref{tab:material_results}]. The exception is object reorientation, where low-friction sensors perform better under vision-only control. In this task, controlled slipping is a desired behavior, and failure modes for high-friction sensors include lifting the object off the table or preventing rotation by applying excessive force.
When tactile feedback is introduced, these failure modes are largely mitigated, and performance across high- and low-friction sensors becomes more comparable. This suggests that high-friction, compliant materials are generally beneficial for manipulation, and that tactile sensing enables policies to avoid the downsides of increased friction by regulating applied forces more precisely.

\textbf{Spatial Resolution:} Across tasks, we observe no significant performance difference between tactile sensors with high spatial resolution (FlexiTac and Daimon) and lower-resolution sensors (FSR, eFlesh, and the contact microphone) when evaluating manipulation success. Prior work has shown that high-resolution visual tactile sensors excel at perception and classification tasks (e.g., contact localization or object recognition) \cite{gao2022objectfolder, schneider2025tactile}; these advantages may not directly translate to improved policy performance in coarse object manipulation. For the tasks studied here, our experiments show that access to task-relevant physical signals is more important than dense spatial representations.

\newpage
\subsection{Sensor Repeatability}
As shown in Figure~\ref{fig:durability_all}, all readings are normalized to the range 0--1, with multidimensional sensors reduced to their maximum value per reading. The FSR is the most consistent across episodes. FlexiTac is consistent when activated, but its resting reading drifts over time. eGain and Daimon perform similarly, exhibiting a ``training phase'' attributable to viscoelastic material behavior. eFlesh drifts consistently, though this may be addressable through changes to the lattice structure. The contact microphone is noisy due to the high-frequency nature of its signal. Daimon exhibits the second lowest variance after the FSR (Fig.~\ref{fig:duribility_daimon}), but this reliability does not consistently translate to higher policy success, suggesting repeatability is not the dominant factor in manipulation performance. Across tasks, open-source and low-cost sensors achieve performance comparable to more expensive or custom alternatives, indicating that accessible tactile hardware can be effective for learning-based manipulation.

We additionally evaluated Daimon, FlexiTac, and eFlesh using flat, round, and wedge tips under push loading at moderate velocity and impact loading at maximum velocity. We report the mean $\pm$ standard deviation of per-episode average tactile readings during contact and no-contact periods (Table~\ref{tab:expanded_repeatability}).

%% file: tables/force_components.tex
\begin{table}[!hb]
\centering
\small
\setlength{\abovecaptionskip}{2pt}
\caption{Reorientation success rates for shear-only and normal-only force inputs.}
\label{tab:force_components}
\begin{tabularx}{\linewidth}{@{}l *{2}{Y}@{}}
\toprule
\textbf{Sensor} & \textbf{Shear only} & \textbf{Normal only} \\
\midrule
Daimon & 0.7 & 0.4 \\
eFlesh & 0.9 & 0.7 \\
\bottomrule
\end{tabularx}
\end{table}

%% file: tables/sensor_material.tex
\begin{table}[!b]
\centering
\small
\setlength{\abovecaptionskip}{2pt}
\setlength{\tabcolsep}{3pt}
\caption{Success rates of vision-only policies grouped by sensor friction.}
\begin{tabularx}{\linewidth}{@{}L{1.2cm} *{3}{Y}@{}}
\toprule
\textbf{Friction Type} & \textbf{\makecell{Pick-and-\\Place}} & \textbf{Insertion} & \textbf{\makecell{Reorien-\\tation}} \\
\midrule
Low       & 0.625 & 0.1 & 0.6 \\
High       & 0.81 & 0.25 & 0.35 \\
\bottomrule
\end{tabularx}
\label{tab:material_results}
\end{table}

%% file: tables/expanded_repeatability.tex
\begin{table*}[!t]
\centering
\small
\setlength{\tabcolsep}{4pt}
\setlength{\abovecaptionskip}{2pt}
\renewcommand{\arraystretch}{1.1}
\caption{Repeatability across tip shapes and loading conditions. Push: moderate velocity; impact: maximum velocity.}
\label{tab:expanded_repeatability}
\begin{tabularx}{\textwidth}{@{}l l *{6}{Y}@{}}
\toprule
& & \multicolumn{2}{c}{\textbf{Daimon}} & \multicolumn{2}{c}{\textbf{FlexiTac}} & \multicolumn{2}{c}{\textbf{eFlesh}} \\
\cmidrule(lr){3-4}\cmidrule(lr){5-6}\cmidrule(l){7-8}
\textbf{Tip} & & \textbf{Impact} & \textbf{Push} & \textbf{Impact} & \textbf{Push} & \textbf{Impact} & \textbf{Push} \\
\midrule
\multirow{2}{*}{Flat} & Contact & $0.561\pm0.024$ & $0.522\pm0.024$ & $0.874\pm0.009$ & $0.828\pm0.007$ & $0.639\pm0.010$ & $0.556\pm0.008$ \\
& No contact & $0.099\pm0.010$ & $0.109\pm0.010$ & $0.277\pm0.156$ & $0.466\pm0.061$ & $0.364\pm0.021$ & $0.281\pm0.014$ \\
\midrule
\multirow{2}{*}{Round} & Contact & $0.259\pm0.005$ & $0.240\pm0.008$ & $0.894\pm0.008$ & $0.859\pm0.009$ & $0.648\pm0.003$ & $0.592\pm0.005$ \\
& No contact & $0.040\pm0.005$ & $0.052\pm0.005$ & $0.265\pm0.029$ & $0.231\pm0.014$ & $0.325\pm0.024$ & $0.274\pm0.012$ \\
\midrule
\multirow{2}{*}{Wedge} & Contact & $0.382\pm0.008$ & $0.372\pm0.011$ & $0.891\pm0.008$ & $0.853\pm0.021$ & $0.683\pm0.012$ & $0.593\pm0.006$ \\
& No contact & $0.059\pm0.007$ & $0.078\pm0.007$ & $0.270\pm0.036$ & $0.298\pm0.183$ & $0.246\pm0.143$ & $0.326\pm0.028$ \\
\bottomrule
\end{tabularx}
\end{table*}

%% file: figures/repeatability_results.tex
\begin{figure*}[tb]
\captionsetup[subfloat]{farskip=4pt}
\centering
\subfloat[FSR\label{fig:duribility_fsr}]{\includegraphics[width=0.32\textwidth]{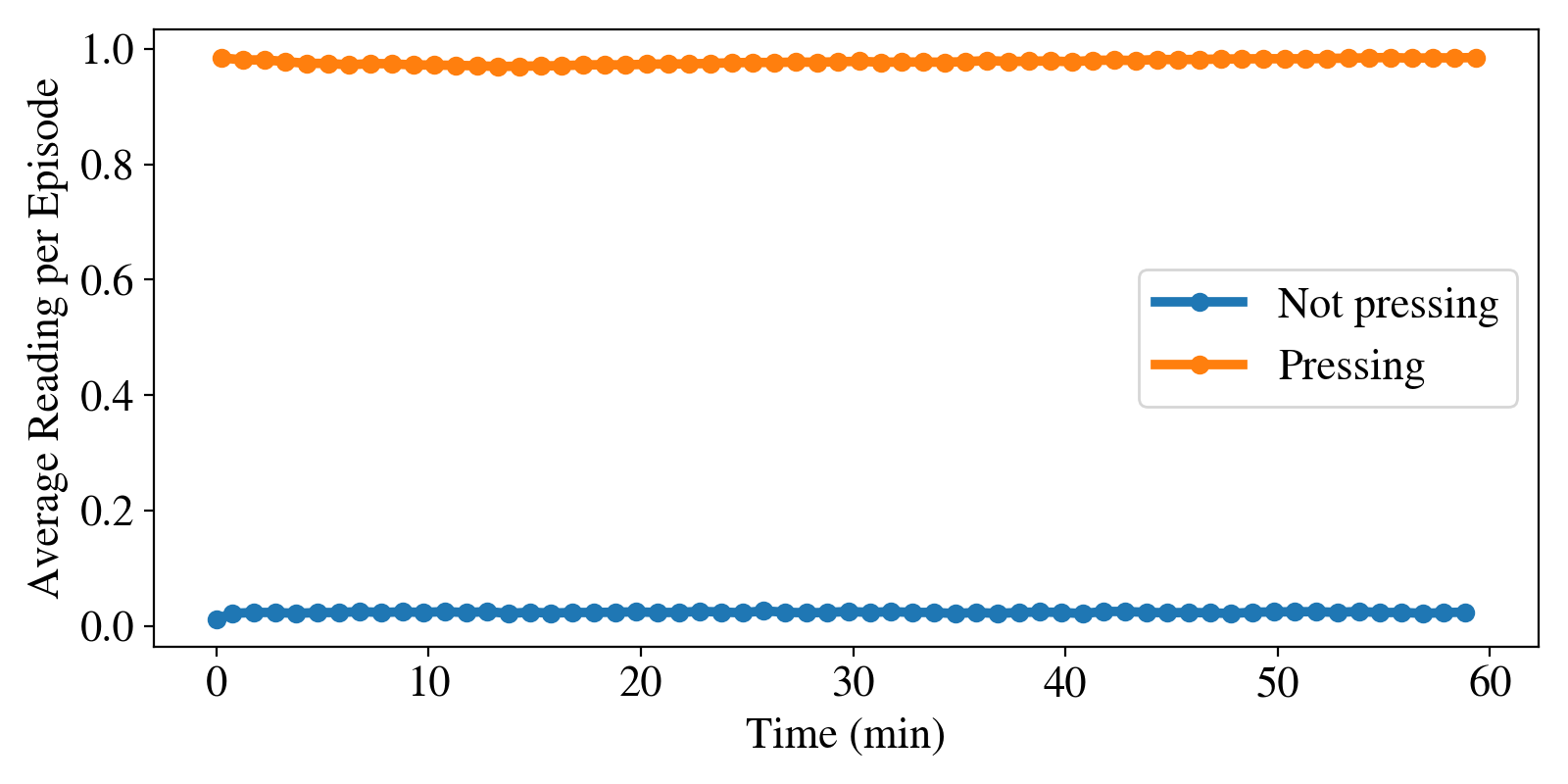}}
\subfloat[FlexiTac\label{fig:duribility_flexitac}]{\includegraphics[width=0.32\textwidth]{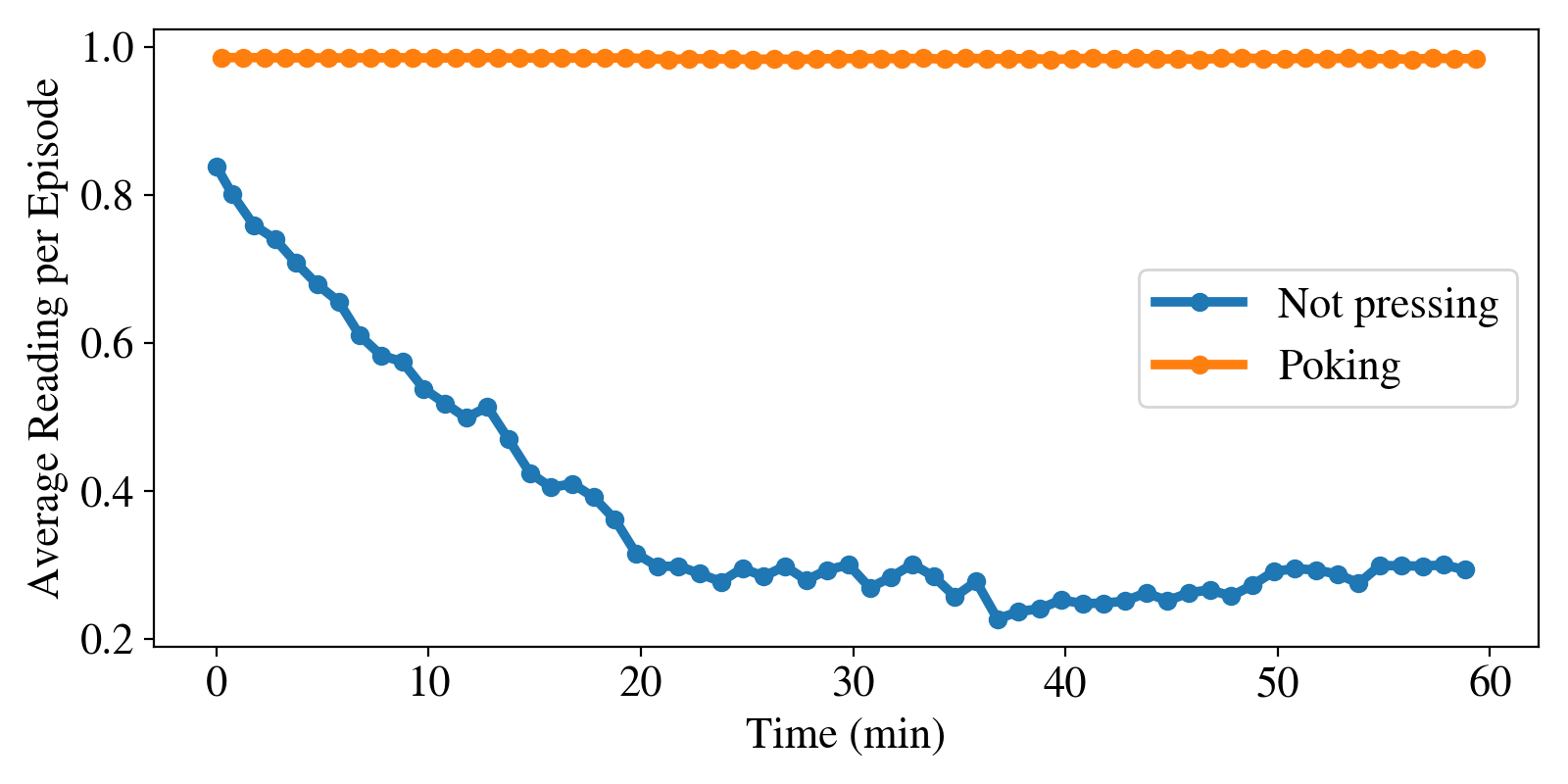}}
\subfloat[eGain\label{fig:duribility_egain}]{\includegraphics[width=0.32\textwidth]{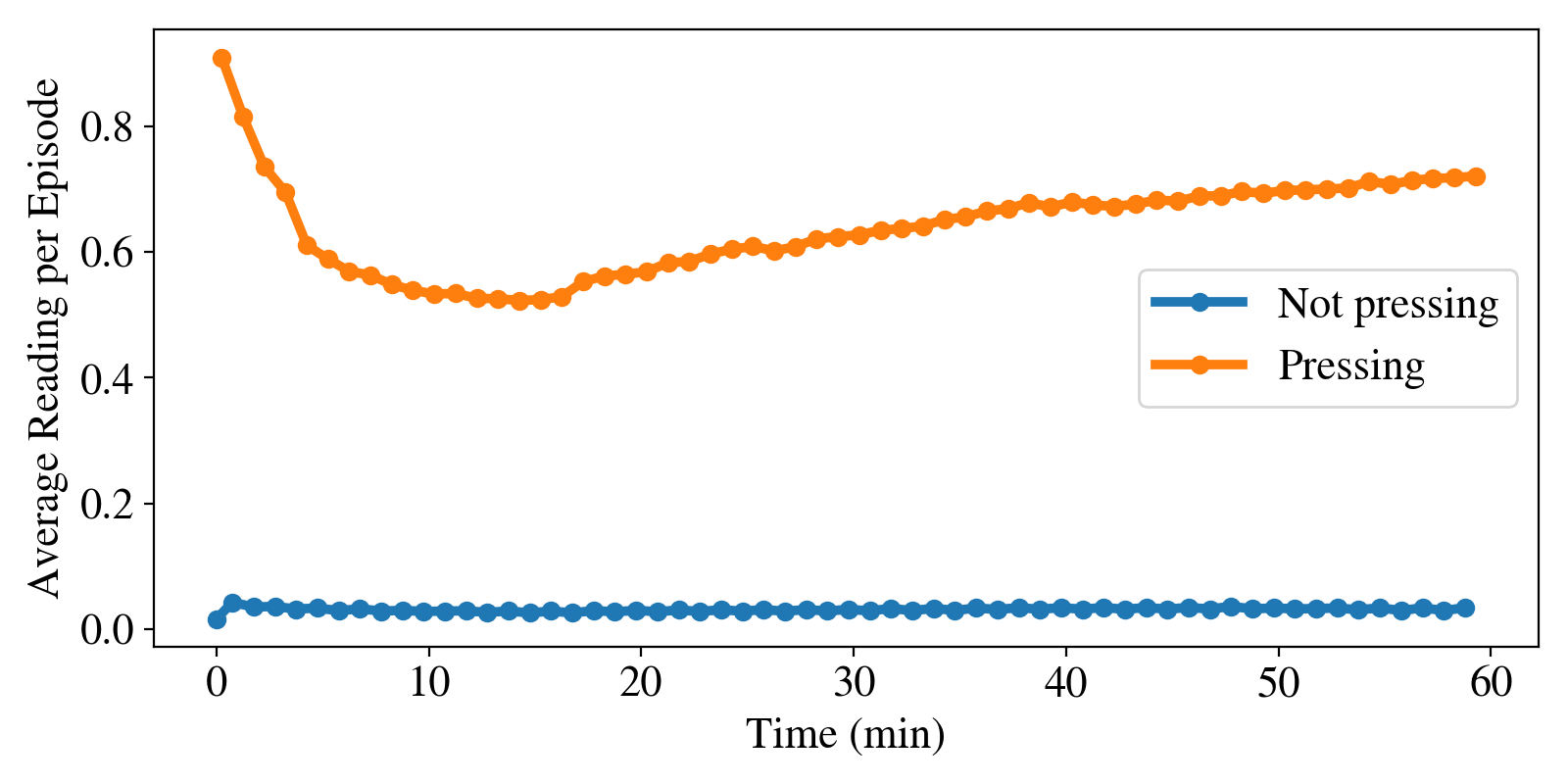}}
\par
\subfloat[eFlesh\label{fig:duribility_eflesh}]{\includegraphics[width=0.32\textwidth]{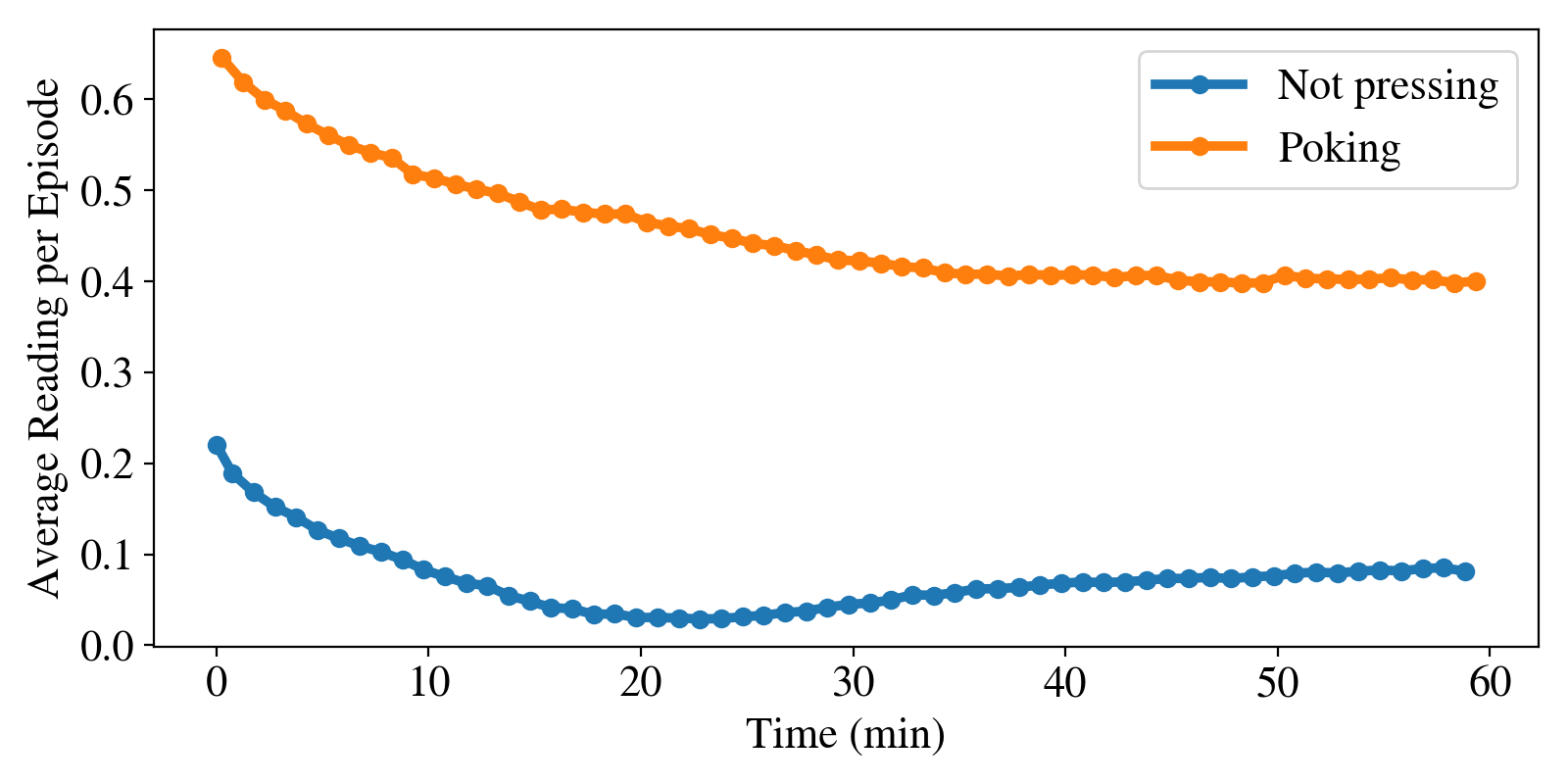}}
\subfloat[Daimon\label{fig:duribility_daimon}]{\includegraphics[width=0.32\textwidth]{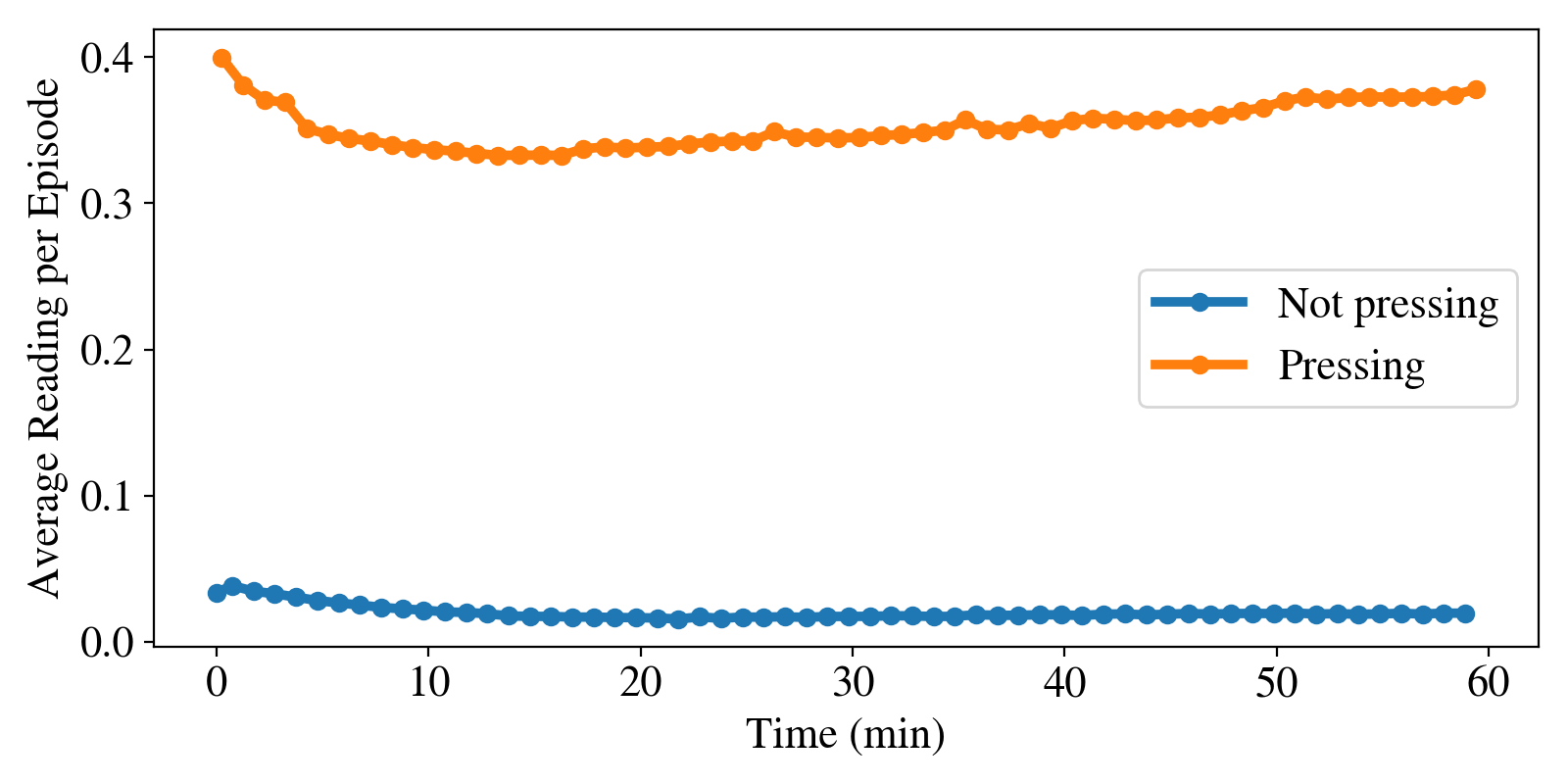}}
\subfloat[Contact Microphone\label{fig:duribility_contact_mic}]{\includegraphics[width=0.32\textwidth]{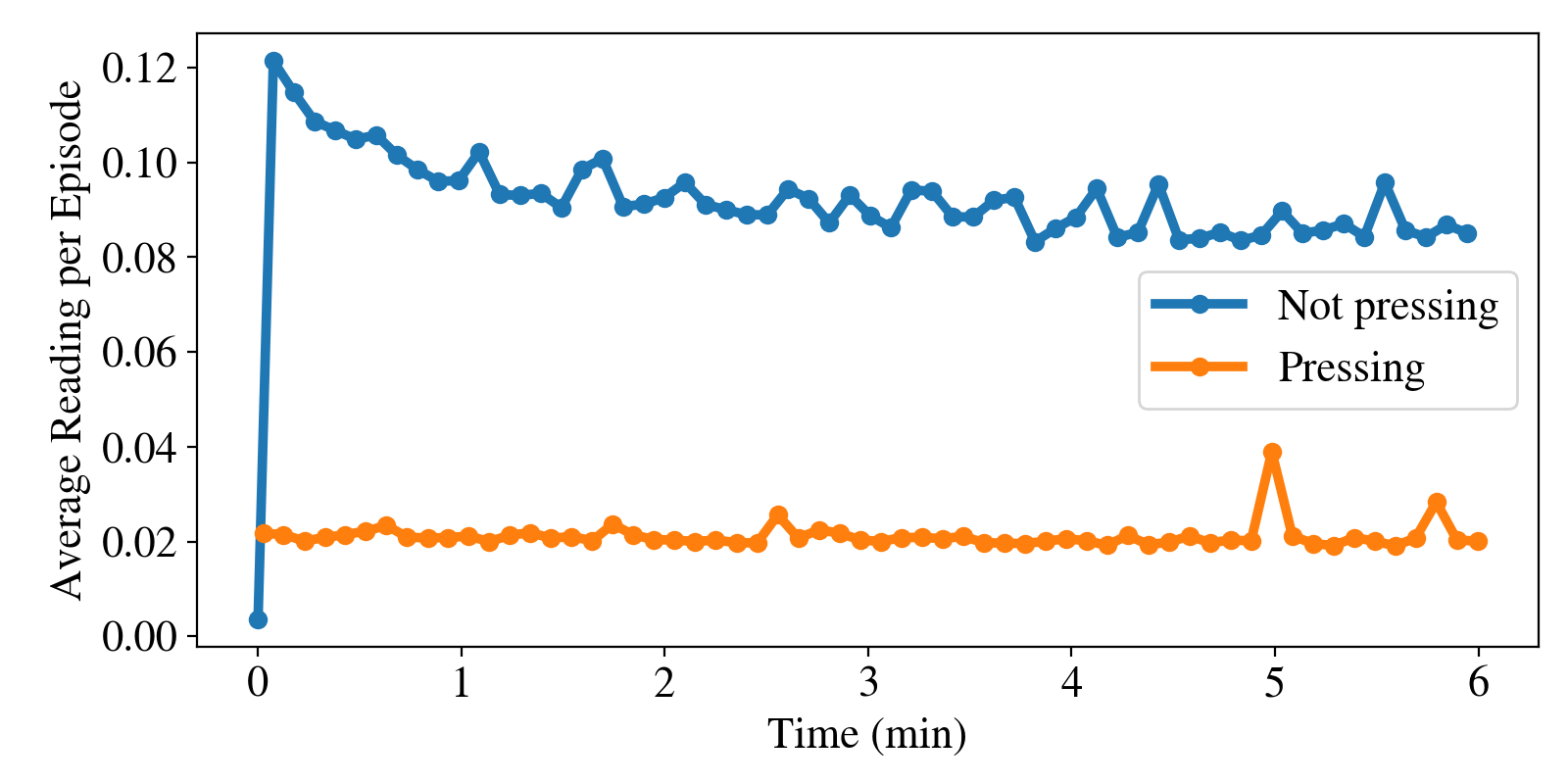}}
\caption{The average sensor reading across episodes during the repeatability test.}
\label{fig:durability_all}
\end{figure*}

%% file: 7_discussion.tex
\section{Conclusion and Limitations}
TacO is a task-driven evaluation of tactile sensors for learning-based object manipulation using a modular imitation learning pipeline. Across multiple manipulation tasks and sensing modalities, we show that incorporating tactile feedback consistently improves policy performance relative to vision-only baselines, and that accessible, open-source sensors can achieve performance comparable to more expensive commercial systems. Sensor material and surface properties impact performance, with high-friction, compliant interfaces better supporting effective tactile interaction. All sensors, code, data, and the repeatability testing apparatus will be released publicly.

This study does not isolate the effect of spatial resolution in tasks requiring fine-grained tactile feedback; future benchmarks targeting dexterous tasks may better reveal the benefits of dense sensing. Additionally, most tasks retain vision throughout execution, and more controlled experiments selectively removing vision during tactile-critical phases could clarify the relative roles of the modalities. Finally, all policies are trained using imitation learning, and evaluating alternative policy architectures beyond ACT would provide a more rigorous study of tactile sensing for manipulation.

%% file: acknowledgments.tex
\section*{Acknowledgments}
We thank Binghao Huang and Venkatesh Pattabiraman for valuable discussions and providing samples of tactile sensors.

%% file: a_appendix.tex
\appendix
\subsection{Additional Tactile Encoders}
\label{sec:additional_tactile_encoders}

As sensor data differ fundamentally in dimensionality and spatial structure, a single shared encoder can't be used. Our encoders used in the main paper experiments are matched to each sensor's signal type: ResNet18 for Daimon, MLP for array-based resistive and magnetic sensors, and MLP on mel-spectrogram features for the contact microphone. Encoder choice has little impact when matched to the signal type, but degrades performance when mismatched. We ran experiments using different encoders for Daimon, FlexiTac, and the Contact Mic. For Daimon we evaluated 3 encoding methods on the pick and place and reorientation tasks. These were ResNet18, PCA analysis, and PCA with a MLP on top of it.
For FlexiTac we create a tactile image using the raw array data, then encode it with all the same encoders as Daimon. The results are shown in Table~\ref{tab:alternative_tactile_encoders}.

\input{tables/alternative_tactile_encoders}

Encoder choice has little impact when matched to the signal type, but degrades performance when mismatched. For FlexiTac pick-and-place, switching from MLP to image-based encoders drops performance from 0.85 to 0.45--0.50; image-based encoders are over-parameterized for the $12\times32$ taxel array. For Daimon reorientation, PCA collapses to 0.00 while ResNet achieves 0.70. In all other conditions, encoder choice has minimal impact. Across experiments, changing the original modality-specific encoder either makes no difference or hurts performance, showing that the original choices were appropriate.

%% file: tables/alternative_tactile_encoders.tex
\begin{table}[H]
\centering
\small
\caption{Vision-only baseline and alternative tactile encoders.}
\label{tab:alternative_tactile_encoders}
\begin{tabularx}{\textwidth}{@{}l l *{5}{Y}@{}}
\toprule
\textbf{Sensor} & \textbf{Task} & \textbf{Vision} & \textbf{MLP} & \textbf{ResNet} & \textbf{PCA} & \textbf{PCA+MLP} \\
\midrule
\multirow{2}{*}{FlexiTac} & Pick \& Place & 0.75 & 0.85 & 0.45 & 0.50 & 0.45 \\
 & Insertion & 0.10 & 0.30 & 0.30 & 0.30 & 0.20 \\
\midrule
\multirow{2}{*}{Daimon} & Pick \& Place & 0.95 & --- & 0.80 & 0.85 & 0.85 \\
 & Reorientation & 0.20 & --- & 0.70 & 0.00 & 0.20 \\
\bottomrule
\end{tabularx}
\end{table}